\documentclass[11pt]{article}

\usepackage[left=2cm, right=2cm, top=2cm, bottom=2cm]{geometry}
\usepackage{setspace}
\usepackage{indentfirst}
\usepackage{tabularx}
\usepackage{xcolor}
\usepackage{soul}
\usepackage{booktabs}        
\usepackage{threeparttable}
\usepackage{natbib}

\sethlcolor{cyan!20} 

\usepackage[T1]{fontenc}
\usepackage{graphicx}
\usepackage{titlesec}
\usepackage{titling}

\usepackage{amsmath}
\usepackage{amssymb}
\usepackage{booktabs}
\usepackage{makecell}
\usepackage{multirow}
\usepackage{longtable}
\usepackage{array}
\usepackage{enumitem}

\usepackage[hyphens]{url}
\usepackage{hyperref}
\hypersetup{
    colorlinks=true,
    linkcolor=blue,
    urlcolor=blue,
    breaklinks=true
}

\pretitle{%
  \begin{center}%
  \rule{\linewidth}{0.6pt}\\[1.5em]
  \LARGE\bfseries
}
\posttitle{%
  \\[0.75em]
  \rule{\linewidth}{0.6pt}
  \end{center}
  \vspace{1.5em}
}

\preauthor{\begin{center}\large}
\postauthor{\end{center}\vspace{0pt}}

\predate{}
\postdate{}
\date{}  

\begin{document}

\title{Heard or Halted?\\[0.5em]
\Large Gender, Interruptions, and Emotional Tone in U.S. Supreme Court Oral Arguments}

\author{
\large Yifei Tong\\[0.25em]
\normalsize Georgetown University\\[0.25em]
\normalsize\texttt{yt583@georgetown.edu}
}

\maketitle

\vspace{-15pt} %
\begin{abstract}
This study examines how interruptions during U.S. Supreme Court oral arguments shape both the semantic content and emotional tone of advocates’ speech, with a focus on gendered dynamics in judicial discourse. Using the ConvoKit Supreme Court Corpus (2010–2019), we analyze 12,663 speech chunks from advocate–justice interactions to assess whether interruptions alter the meaning of an advocate’s argument and whether interruptions toward female advocates exhibit more negative emotional valence.

Semantic shifts are quantified using GloVe-based sentence embeddings, while sentiment is measured through lexicon-based analysis. We find that semantic similarity between pre- and post-interruption speech remains consistently high, suggesting that interruptions do not substantially alter argumentative content. However, interruptions directed at female advocates contain significantly higher levels of negative sentiment. These results deepen empirical understanding of gendered communication in elite institutional settings and demonstrate the value of computational linguistic methods for studying power, discourse, and equity in judicial proceedings.
\end{abstract}


\section{Introduction}  
\vspace{-5pt} %

Oral arguments before the U.S. Supreme Court are a central component of judicial decision making and one of the few public opportunities to observe how justices and advocates interact. Interruptions in this setting can signal disagreement, redirect the flow of argument, or reflect underlying power relations within the Court. Prior research shows that female advocates and female justices are interrupted more frequently than their male counterparts, even when accounting for factors such as seniority and experience \citep{Cai2025, JacobiSchweers2017}. These findings demonstrate that interruptions are not distributed evenly across participants and that gender plays a meaningful role in shaping communicative dynamics.

Less understood are the consequences of these interruptions for the substance and tone of advocates' speech. While the frequency of interruptions has been well documented, it remains unclear whether interruptions alter the semantic content of arguments or whether the emotional tone of interruptions differs by gender. This study mainly examines two questions:

\begin{enumerate}
\item Do interruptions alter or preserve the core semantic content of an advocate's argument?
\item Are interruptions directed at female advocates characterized by more negative sentiment than those directed at male advocates?
\end{enumerate}

Understanding these effects is important for evaluating how communicative practices influence advocacy, credibility, and courtroom equity. If interruptions meaningfully shift semantic content, they may disrupt argument structure or reduce an advocate's ability to present a coherent position. If interruptions toward women exhibit systematically more negative sentiment, this may indicate gendered asymmetries in how advocates are treated during oral argument.

To investigate these questions, we combine computational techniques with institutional context. Word embeddings allow us to measure semantic similarity before and after interruptions, and lexicon-based sentiment analysis provides a way to assess emotional tone. By applying these methods to a decade of Supreme Court oral argument transcripts, we move beyond documenting interruption frequency to analyze how interruptions function within the broader communicative structure of the Court.

\section{Data}
\vspace{-5pt} %
\subsection{Data Source}

This study draws on the ConvoKit Supreme Court Corpus, a structured collection of oral argument transcripts from the United States Supreme Court. Transcript structure and case-level metadata were obtained from the Supreme Court Database \citep{Spaeth2021} and processed using components of the ConvoKit toolkit \citep{ConvoKit2019}. We focus on the 2010–2019 terms, which provide consistent metadata and high-quality transcript formatting. After filtering, our final analytic sample includes 12,663 speech chunks produced by legal advocates.

A speech chunk is defined as a continuous segment of an advocate’s speech during an interaction with a single justice. Chunks are constructed by grouping together consecutive advocate utterances that occur between interruptions or natural pauses in the exchange. Each chunk, therefore, represents a coherent portion of an advocate’s argument, typically spanning several sentences, and serves as the primary unit of observation in our analysis.

Each utterance in the corpus includes detailed metadata, including the case identifier, advocate name, speaker role, and the full transcript text. These attributes allow us to track interaction sequences between advocates and justices and to link segments of speech to specific cases. Advocate gender is inferred using WGND 2.0, a curated name–gender reference dataset hosted on Harvard Dataverse \citep{Raffo2021}. This resource provides standardized mappings between first names and gender categories and is widely used in computational social science for demographic inference. Using this dataset ensures consistent classification of advocates as male or female across the corpus. The selected time range provides the most complete coverage of these metadata fields, which is essential for reliable extraction of interactional and demographic features.

Our data filtering procedures follow the logic outlined in original literature \citep{Cai2025}. We exclude transcripts with incomplete metadata, turns in which the advocate cannot be reliably identified, and cases where the subject matter is likely to introduce strong topic-driven sentiment. In particular, we remove cases tagged with a \texttt{female\_issue} indicator, such as abortion, to avoid confounding emotional language associated with highly sensitive issue areas. These topics routinely elicit strong affective responses from both justices and advocates, which may overshadow ordinary interactional patterns and inflate sentiment measures in ways unrelated to gender dynamics. Although we adapt the original implementation for our own environment, the filtering criteria and underlying rationale remain consistent with the authors’ approach.

\subsection{Feature Construction}

Interruptions were programmatically identified by detecting instances in which an advocate’s utterance was prematurely cut off and immediately followed by a justice’s interjection. In the transcript data, such interruptions are typically marked by double dashes (“--”) to indicate an abrupt cutoff or by ellipses (“…”) to signal trailing or incomplete speech. These textual markers were cross-referenced with speaker metadata to ensure that the following turn belonged to a justice, allowing us to isolate justice-initiated interruptions with high precision.

The exchange below illustrates a typical interruption event:

\begin{quote}
\textbf{Douglas Laycock (Advocate):\\}
\textit{“Well, some courts have said yes. There's very little in this record about full beards and whether they're safe or whether they're dangerous…”}

\textbf{Justice Scalia (Interrupting):\\}
\textit{“Mr. Laycock, the problem I have with— with your client's claim…”}
\end{quote}

Here the justice interjects while the advocate is still developing his point, elaborating on how the evidentiary record relates to the broader legal question. The interruption cuts off this explanation mid-sentence and redirects the conversation toward the justice’s preferred line of inquiry. In our dataset, such events are detected at the utterance level and used to determine whether a speech chunk is classified as interrupted. Specifically, any speech chunk containing an interruption marker is assigned a value of \texttt{interrupted = 1}, while all others receive \texttt{0}. This binary indicator serves as a central feature in the analyses that follow, structuring both the semantic similarity comparisons and the sentiment evaluation.

\section{Methodology}
\vspace{-5pt} %
\subsection{Sentiment Similarity Analysis}

To evaluate whether interruptions alter the semantic content of an advocate’s argument, we measure semantic similarity between interrupted and uninterrupted speech using pre-trained GloVe word embeddings. Embedding-based approaches are widely used to represent semantic meaning in legal and political texts because they capture distributional relationships between words in continuous vector space \citep{Mikolov2013, Pennington2014}.

All chunks were first preprocessed by lowercasing text, removing punctuation, expanding contractions, and removing tokens not present in the GloVe vocabulary. For each remaining token, we retrieved its corresponding 100-dimensional GloVe vector. To illustrate, consider the example chunk:

\begin{quote}
\textit{“\hl{Religious} \hl{belief} is \hl{personal}, but \hl{prison} \hl{policy} must be \hl{reasonable}.”}
\end{quote}

After preprocessing, each word in the sentence maps to a 100-dimensional embedding, producing an \textit{N} × 100 matrix where \textit{N} is the number of valid tokens. In the example chunk, we highlight several tokens (such as religious, belief, and policy) to illustrate words that remain after preprocessing and appear in the GloVe vocabulary. Each highlighted word corresponds to one row in the embedding matrix. Averaging across the rows yields a single 100-dimensional sentence vector that summarizes the semantic meaning of the entire chunk. We use the 100-dimensional version of GloVe because it provides a strong balance between representational richness and computational efficiency. Higher-dimensional vectors often provide diminishing gains in similarity tasks while substantially increasing memory and computational cost \citep{Pennington2014}.

To assess whether interruptions correspond to shifts in meaning, we compare embeddings within each advocate. For every advocate, we construct two composite embeddings: one representing all interrupted chunks and another representing all uninterrupted chunks. Cosine similarity between these vectors provides a measure of semantic overlap, indicating how closely the average meaning of interrupted speech aligns with an advocate’s typical argumentative content. Cosine similarity values near 1 indicate substantial preservation of meaning, whereas notably lower values suggest potential semantic divergence between interrupted and uninterrupted speech. In practice, values above approximately 0.85 typically reflect strong semantic stability, and we use this threshold as a guiding benchmark when interpreting the results.

This embedding-based procedure yields a compact and interpretable measure of semantic continuity and allows us to test whether interruptions are associated with changes in what advocates say rather than simply how they are allowed to say it. These similarity scores form the basis of the semantic analysis presented in Section 4.

\subsection{Sentiment Analysis}

To evaluate whether interruptions directed at female advocates carry more negative emotional tone than those directed at male advocates, we apply lexicon-based sentiment analysis to the subset of speech chunks identified as interrupted. Emotional valence is quantified using the NRC Emotion Lexicon \citep{MohammadTurney2013}, a widely used resource in computational linguistics that assigns words to discrete affective categories, including positive and negative sentiment as well as specific emotions such as anger, fear, sadness, and disgust. The NRC lexicon is particularly suitable for this task because it provides fine-grained emotion labels that have been validated across multiple domains, and it performs reliably in short, argumentative text where explicit affective markers are sparse but meaningful.

For each interrupted chunk, we count the number of tokens associated with NRC’s selected emotion categories: positive, negative, anger, fear, sadness, and disgust. We then compute an aggregate measure of negativity, referred to as the negative ratio, which is defined as the proportion of negative-affect words among all emotion-bearing words in the chunk:

\[
\text{neg\_ratio} =
\frac{
\text{negative} + \text{anger} + \text{fear} + \text{sadness} + \text{disgust}
}{
\text{total emotion words}
}
\]

This ratio normalizes emotional expression by the total number of emotion-related tokens, providing a scale-free measure that allows for comparisons across advocates and across gender groups despite substantial variation in chunk length.

We then compare the distribution of negative ratios between male and female advocates to assess whether interruptions toward women exhibit systematically more negative affect. This comparison forms the basis for the sentiment results presented in Section 4.

\section{Results}
\vspace{-5pt} %
\subsection{Semantic Shifts: Sentiment Similarity Analysis}

\begin{figure}[ht]
\centering
\includegraphics[width=14cm]{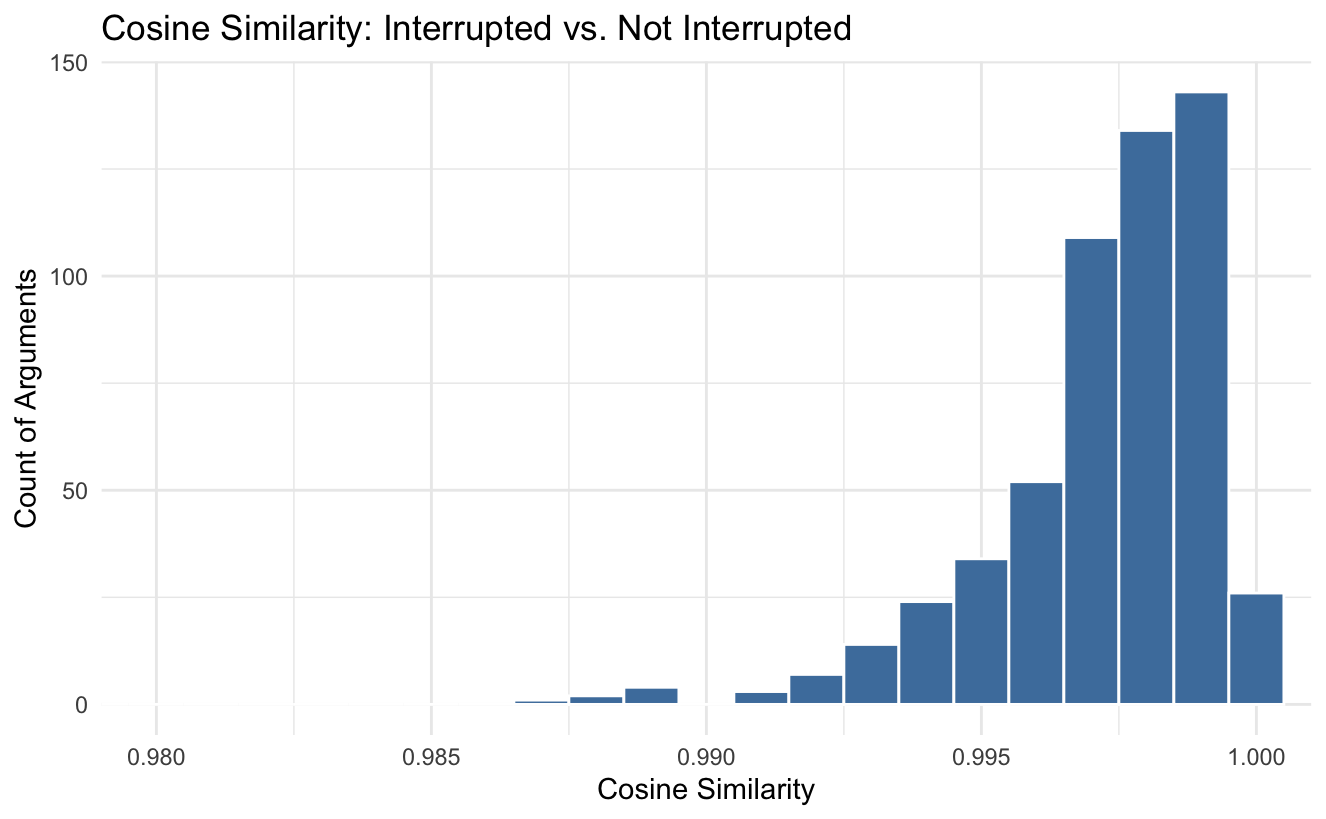}
\caption{Distribution of Semantic Overlap Between Interrupted and Uninterrupted Speech}
\label{fig:cosine}
\end{figure}

\newpage
To evaluate whether interruptions alter the semantic content of an advocate’s argument, we computed cosine similarity between the average embeddings of interrupted and uninterrupted chunks for each advocate. These embeddings represent the underlying semantic structure of each chunk, which allows us to assess whether interruptions correspond to meaningful shifts in argumentative content.

Figure~\ref{fig:cosine} displays the distribution of cosine similarity scores. The scores are highly concentrated between 0.98 and 1.00, with a strong peak above 0.99. Table~\ref{tab:cos_summary} reports the corresponding summary statistics. The mean similarity is 0.997, and the median is 0.9977, indicating an exceptionally high degree of semantic overlap between interrupted and uninterrupted chunks produced by the same advocate.

\begin{table}[htbp]
\centering
\caption{Summary Statistics of Cosine Similarity Scores}
\label{tab:cos_summary}
\begin{tabular}{cccccc}
\toprule
\textbf{Min} & \textbf{1st Qu.} & \textbf{Median} & \textbf{Mean} & \textbf{3rd Qu.} & \textbf{Max} \\ 
\midrule
0.8899 & 0.9965 & 0.9977 & 0.9970 & 0.9986 & 0.9999 \\ 
\bottomrule
\end{tabular}
\end{table}

The near-perfect similarity scores suggest that interruptions do not substantially alter the semantic content of advocates’ arguments, at least as measured by GloVe-based embeddings. Rather than shifting meaning, interruptions appear to influence other aspects of communication such as rhythm, perceived control of the conversational floor, or patterns of interaction. Advocates typically return to their original argumentative trajectory after being interrupted, which helps maintain semantic consistency despite temporary breaks in delivery. This pattern of results is consistent with prior work that characterizes interruptions in judicial and political discourse as indicators of power dynamics or interactional dominance. In this context, the high level of semantic stability supports the view that interruptions function primarily as conversational interventions rather than catalysts for meaning change in argumentative content.

\subsection{Gendered Sentiment: Lexicon-Based Analysis}
\subsubsection{Descriptive Patterns}

\begin{figure}[ht]
\centering
\includegraphics[width=15cm]{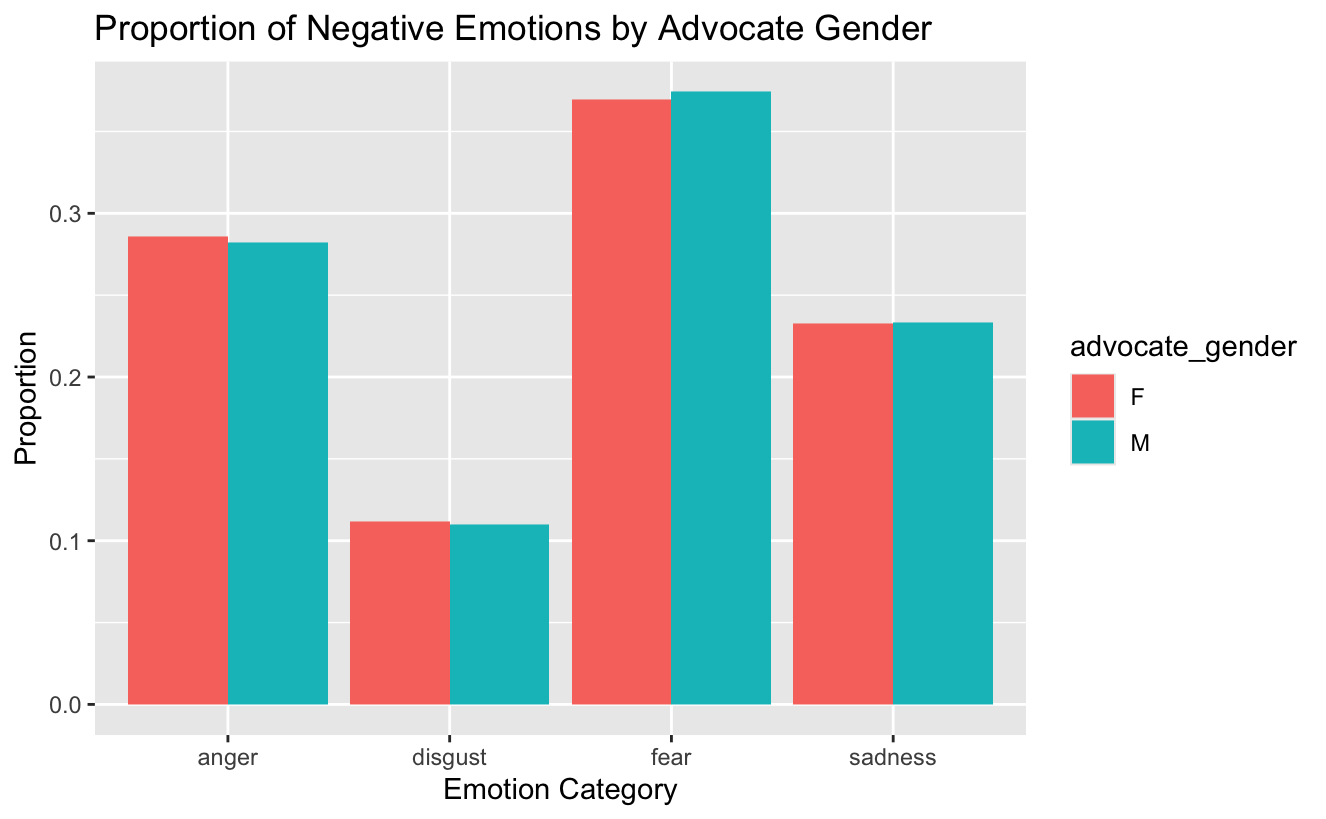}
\caption{Proportion of Negative Emotions in Interrupted Chunks by Advocate Gender}
\label{fig:proportion}
\end{figure}

To assess whether interruptions directed at female advocates exhibit more negative emotional tone, we applied lexicon-based sentiment analysis using the NRC Emotion Lexicon to all interrupted chunks. Each chunk was converted into counts of emotion-bearing words, and a negative sentiment ratio was calculated as the share of negative-affect tokens among all emotion-labeled tokens.

Figure~\ref{fig:proportion} presents the distribution of negative emotion categories across advocate gender. The proportions of anger, disgust, fear, and sadness are broadly similar for male- and female-directed interruptions, with nearly overlapping values across all four categories. Female advocates show slightly higher proportions of anger and fear, while male advocates exhibit marginally higher levels of sadness, but these differences remain small and fall within a relatively narrow range. The descriptive evidence therefore suggests that gender differences are unlikely to arise from systematic variation in specific negative emotions, and are instead more likely to reflect modest shifts in the overall intensity of negative sentiment.

\begin{figure}[ht]
\centering
\includegraphics[width=15cm]{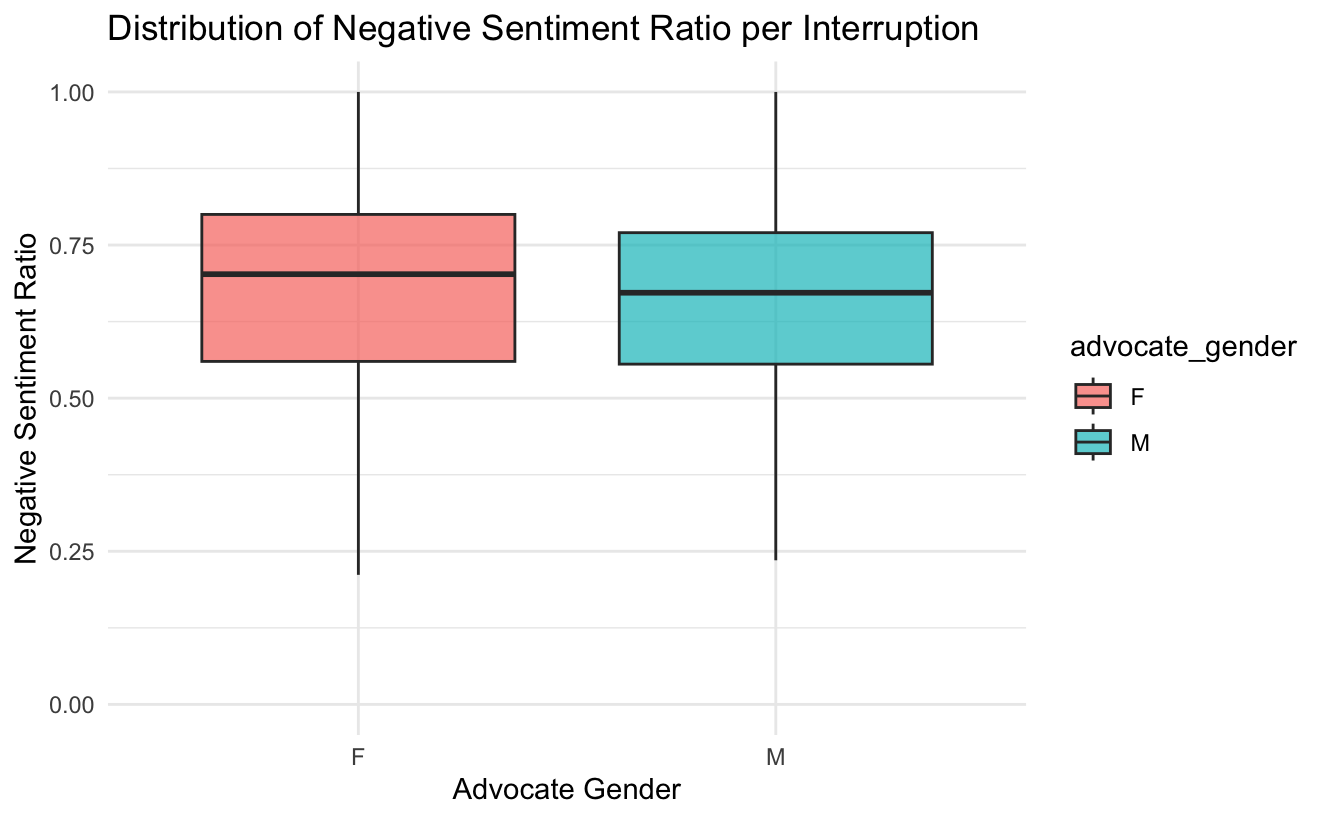}
\caption{Distribution of Negative Sentiment Ratios in Interrupted Chunks by Advocate Gender}
\label{fig:neg_ratio}
\end{figure}

Figure~\ref{fig:neg_ratio} displays the distribution of the negative sentiment ratio for each interruption. Female advocates exhibit a modest upward shift in both the median and upper quartile relative to male advocates, indicating a somewhat higher concentration of negative sentiment in their interrupted speech. Although the distributions overlap substantially, the shift is consistent with the possibility that interruptions directed at women carry slightly more negative emotional tone. These descriptive patterns suggest a small but visible gender difference that warrants formal statistical evaluation.

\subsubsection{Statistical Evidence}

To evaluate whether the descriptive differences in negative sentiment are statistically meaningful, we conducted a Welch two-sample t-test comparing the negative sentiment ratio for interruptions directed at male and female advocates. The test indicates a statistically significant difference ($p = 0.0078$), with interruptions directed at female advocates exhibiting, on average, a higher negative sentiment ratio. The estimated difference in means is approximately two percentage points, and the 95 percent confidence interval [0.0039, 0.0256] does not include zero. Although the magnitude is modest, the confidence interval indicates that the observed difference is unlikely to have arisen by chance.

To assess whether this pattern persists after accounting for potential confounding factors, we estimated an ordinary least squares regression with the negative sentiment ratio as the dependent variable. The model includes controls for advocate experience, case year, justice–advocate ideological alignment, and advocate ideology. Table~\ref{tab:sentiment_regression} reports the key estimates. The coefficient for male advocates is negative and statistically significant ($p = 0.0298$), which indicates that interruptions directed at female advocates contain a higher proportion of negative sentiment even after adjusting for covariates. Advocate experience is also associated with lower negativity, consistent with the idea that experienced advocates may navigate interruptions with greater composure or rhetorical control.

The model’s explanatory power is small, which is expected for fine-grained linguistic variation in short legal utterances, yet the consistency of the gender coefficient across both the t-test and regression provides evidence that the effect is meaningful. These findings reinforce the descriptive patterns and support the conclusion that interruptions directed at female advocates carry slightly more negative emotional tone.

\begin{table}[htbp]
\centering
\begin{threeparttable}
\caption{Predicting Negative Sentiment Ratio in Interruptions}
\label{tab:sentiment_regression} 
\begin{tabular}{lccc}
\toprule
\textbf{Variable} & \textbf{Estimate} & \textbf{Std. Error} & \textbf{p-value} \\
\midrule
(Intercept) & -1.4088 & 1.2816 & 0.2717 \\
Advocate gender (Male) & -0.0114* & 0.0052 & 0.0298 \\
Advocate experience (int) & -0.0022*** & 0.0004 & $< 0.001$ \\
Case year & 0.0010 & 0.0006 & 0.1064 \\
Female issue\tnote{a} & --- & --- & --- \\
Advocate ideology (Liberal) & 0.0061\tnote{\textdagger} & 0.0035 & 0.0861 \\
Ideology matches & 0.0018 & 0.0036 & 0.6196 \\
\midrule
Residual Std. Error & \multicolumn{3}{l}{0.1785 (df = 10291)} \\
Multiple $R^2$ & \multicolumn{3}{l}{0.0041} \\
Adjusted $R^2$ & \multicolumn{3}{l}{0.0036} \\
F-statistic & \multicolumn{3}{l}{8.501 on 5 and 10291 DF, $p < 0.001$} \\
\bottomrule
\end{tabular}

\begin{tablenotes}
\footnotesize
\item[*] $p < 0.05$ \quad *** $p < 0.001$ \quad \textsuperscript{\textdagger} $p < 0.1$
\item[a] Variable dropped due to collinearity or lack of variance.
\end{tablenotes}
\end{threeparttable}
\end{table}

\subsubsection{Interpretation}
Taken together, the descriptive and statistical analyses indicate a small but consistent gender difference in the emotional tone of interruptions. Although the distributions overlap heavily and the effect size is modest, interruptions directed at female advocates contain a slightly higher proportion of negative emotional language. This pattern persists after accounting for experience, case characteristics, and ideological factors. Overall, the evidence suggests that gendered dynamics in Supreme Court oral arguments extend beyond the frequency of interruptions to include subtle qualitative differences in how those interruptions unfold.

\subsection{LDA for Topic Modeling}

To explore whether gender differences in emotional tone might reflect underlying variation in the subject matter of advocate speech, we implemented Latent Dirichlet Allocation (LDA) topic modeling. LDA is frequently used in studies of political and legal discourse to uncover latent thematic structure in text corpora, making it a natural candidate for assessing whether systematic topic differences might explain variation in sentiment across genders \citep{GrimmerStewart2013}. We fit a six-topic model using a document-term matrix constructed from cleaned chunk text, assigning each chunk a probability distribution across topics.\footnote{Appendix Figures display, respectively, the top terms associated with each topic and the distribution of average topic weights by advocate gender.}

As a robustness check, we compared topic distributions across gender and also included topic weights as covariates in regression models predicting negative sentiment. Average topic proportions were nearly identical for male and female advocates, and the coefficient on gender remained significant after adjusting for topic weights. These results indicate that topic content does not account for the gender-based disparities in emotional tone documented earlier.

Inspection of the top terms across the six topics further illustrates the limitations of applying LDA in this domain. High-probability tokens showed substantial lexical overlap across topics, dominated by standard legal terminology such as court, justice, law, and statute. This dominance of common legal vocabulary is consistent with the highly structured nature of Supreme Court argumentation and limits LDA’s ability to distinguish meaningful thematic clusters. More broadly, these patterns reflect several structural features of the corpus:

\begin{itemize}
\item \textbf{Highly constrained vocabulary:} Legal discourse is formal and repetitive, relying on standardized phrases, citations, and doctrinal language. This uniformity reduces topic heterogeneity and constrains the formation of distinct topics.
\item \textbf{Short, focused documents:} LDA assumes that documents are long enough to exhibit a mixture of topics, yet individual speech chunks tend to be narrow in scope and often respond directly to a justice’s question.

\item \textbf{Limited thematic separation:} The topics extracted by LDA fail to capture substantive differences in legal content and instead reflect small variations of generalized courtroom language or "court talk".

\end{itemize}

In sum, while topic modeling was conceptually appropriate as an exploratory extension, the linguistic constraints of Supreme Court oral arguments combined with the repetitive and domain-specific vocabulary limited its analytical value. We interpret the LDA results primarily as evidence of domain-specific challenges rather than topic-based gender divergence.

\section{Conclusions}
\vspace{-5pt} %

This study examined how interruptions during U.S. Supreme Court oral arguments affect both the meaning and emotional tone of advocates’ speech, with particular focus on gender dynamics. Drawing on a decade of transcripts and more than twelve thousand speech chunks, we evaluated whether interruptions change what advocates say or how they are spoken to. Using GloVe-based embeddings, we found that interrupted and uninterrupted speech from the same advocate exhibits extremely high cosine similarity, indicating that interruptions do not meaningfully alter the semantic content of arguments. Advocates typically resume the same substantive trajectory even after conversational disruptions, suggesting that interruptions operate more as breaks in delivery than as shifts in meaning.

At the same time, sentiment analysis reveals subtle but consistent gender differences in the emotional tone of interruptions. Interrupted speech directed at female advocates contains a modestly higher proportion of negative sentiment compared with that directed at male advocates, a pattern that remains statistically significant after accounting for experience, ideology, and case characteristics. While the magnitude of these differences is small, their consistency suggests that interruptions may reproduce unequal interactional dynamics even within a highly formal and regulated institutional setting. Together, these findings deepen our understanding of how power and communication intersect in elite legal discourse.

\subsection{Theoretical and Empirical Contributions}

This project contributes to research on judicial communication and gendered interaction in several important ways. First, it advances the literature by shifting attention from the frequency of interruptions, which has been the focus of earlier work \citep{Cai2025, JacobiSchweers2017}, to their substantive consequences. By examining both semantic content and emotional tone, the study evaluates whether interruptions act as meaningful interventions in an advocate’s argument or whether they primarily serve as markers of conversational dominance.

Second, the results highlight the need to understand interruptions as rhetorical acts that operate within established institutional norms. The high level of semantic stability suggests that interruptions rarely compel advocates to alter their substantive claims. At the same time, the finding that interruptions directed at female advocates contain slightly higher negative emotional content corresponds with scholarship showing that subtle linguistic cues can reinforce hierarchies and influence perceptions of authority in legal settings. This aligns with sociolegal analyses that describe courtroom discourse as a site where power relations are enacted through interactional practices even when formal procedures appear neutral \citep{Matoesian2001}.

Third, this project demonstrates the potential and current limitations of computational approaches for the study of legal language. Sentence embeddings provide a practical method for assessing semantic consistency, while lexicon-based sentiment analysis offers a window into the emotional tone of judicial interactions. Taken together, these tools form a methodological foundation that can be extended to legislative hearings, lower court proceedings, or other institutional deliberative settings. They also contribute to a growing body of work in political communication, computational social science, and the study of law and courts that seeks to quantify discourse in systematic and interpretable ways.

\subsection{Limitations}
Although this study offers new evidence on the semantic and emotional characteristics of interruptions in Supreme Court oral arguments, several limitations should be noted. First, the semantic analysis relies on static word embeddings. These models assign a single vector to each word and therefore cannot account for contextual nuance, pragmatic meaning, or shifts in argumentative framing. More recent advances in natural language processing, including contextualized models such as BERT or Sentence-BERT, have demonstrated greater sensitivity to sentence-level meaning and would likely detect semantic variation with more precision \citep{Devlin2019, ReimersGurevych2019}.

Second, the sentiment analysis uses a dictionary-based approach, which interprets tokens independently of surrounding context. This method performs less effectively in formal settings where emotional tone is subtle or conveyed indirectly. Recent evaluations of lexicon-based sentiment tools show that they tend to under-identify affective expression in institutional or expert domains \citep{HuttoGilbert2014}. As a result, the emotional patterns we detect may represent conservative estimates of the true variation in rhetorical tone.

Third, interruptions are coded as a binary event, which simplifies the complex dynamics of turn-taking. Some interruptions are brief clarifications, while others redirect an advocate’s reasoning or challenge a central claim. Treating all interruptions as equivalent may obscure important variation in their force, intent, or strategic function. Future research could benefit from distinguishing among types of interruptions and measuring severity or clustering.

Finally, the study focuses exclusively on the U.S. Supreme Court. The interactional norms of this institution are highly formal and shaped by long-established procedural traditions. Research on communication patterns among justices shows that questioning styles and interactional behavior vary meaningfully across individuals, suggesting that institutional context and participant characteristics influence how interruptions unfold \citep{FeldmanGill2019}. These contextual differences may limit the generalizability of the present findings beyond this setting.

\subsection{Future Directions}

Several promising avenues offer opportunities to extend this work. First, future studies could apply contextualized language models to examine whether interruptions lead to more subtle shifts in meaning that static embeddings cannot detect. Models such as Sentence-BERT provide richer sentence-level representations and have demonstrated improved performance in identifying fine-grained semantic differences, particularly in domains that involve complex reasoning or specialized vocabulary \citep{ReimersGurevych2019}. Incorporating these methods may reveal semantic changes that were not captured through GloVe-based averaging.

Second, researchers could incorporate more detailed measures of interruption characteristics. Interruptions vary widely in strength, function, and conversational impact. Some are brief procedural clarifications, while others redirect an advocate’s reasoning, challenge a central claim, or signal disagreement. Modeling features such as interruption duration, timing, degree of overlap, or sequential positioning may allow researchers to better capture the interactional force of interruptions. This would support a more nuanced understanding of how interruptions shape argumentative flow and whether certain types affect advocates differently.

Third, justice-level heterogeneity remains an important direction for further investigation. Justices differ in their questioning style, levels of assertiveness, and patterns of engagement with advocates. A dyadic modeling approach that pairs individual justices with individual advocates could identify whether certain justices systematically direct harsher or more negative interruptions toward women, or whether the observed patterns are broadly consistent across the bench.

Fourth, extending the empirical setting would help assess the generalizability of these findings. The U.S. Supreme Court represents a particularly formal and hierarchical environment, and its communicative norms may differ from those of federal appellate courts, state supreme courts, or international tribunals. Research on judicial behavior indicates that courts vary widely in their interactional norms and questioning styles, suggesting that comparing across institutional contexts may reveal whether the dynamics observed here reflect broader features of adversarial discourse or features specific to the Supreme Court.

Finally, incorporating audio-based features presents a particularly promising direction. Prosodic cues, including pitch, loudness, and timing, carry important information about authority, alignment, and interpersonal stance. Research on prosodic entrainment shows that these cues can reflect subtle interactional dynamics that are not captured by transcripts alone \citep{LuboldPonBarry2014}. Combining text-based and audio-based measures may therefore provide a more comprehensive account of how interruptions shape the emotional and relational dimensions of legal discourse.

\newpage
\section*{Appendix}

\begin{figure}[!ht]
    \centering
    \includegraphics[width=15cm]{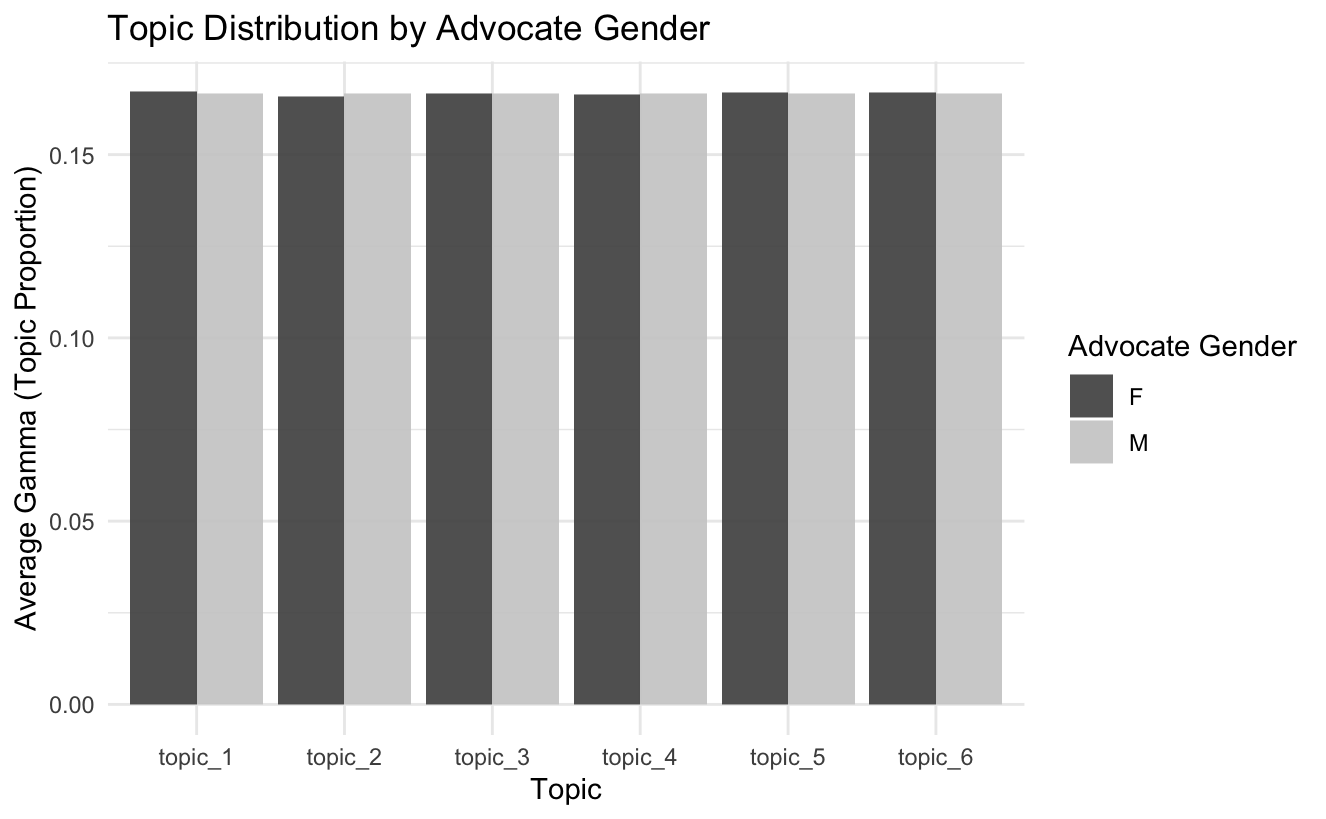}
    \caption{LDA Topic Distribution by Advocate Gender}
    \label{fig:lda_topic_gender}

    \vspace{1cm} 

    \includegraphics[width=15cm]{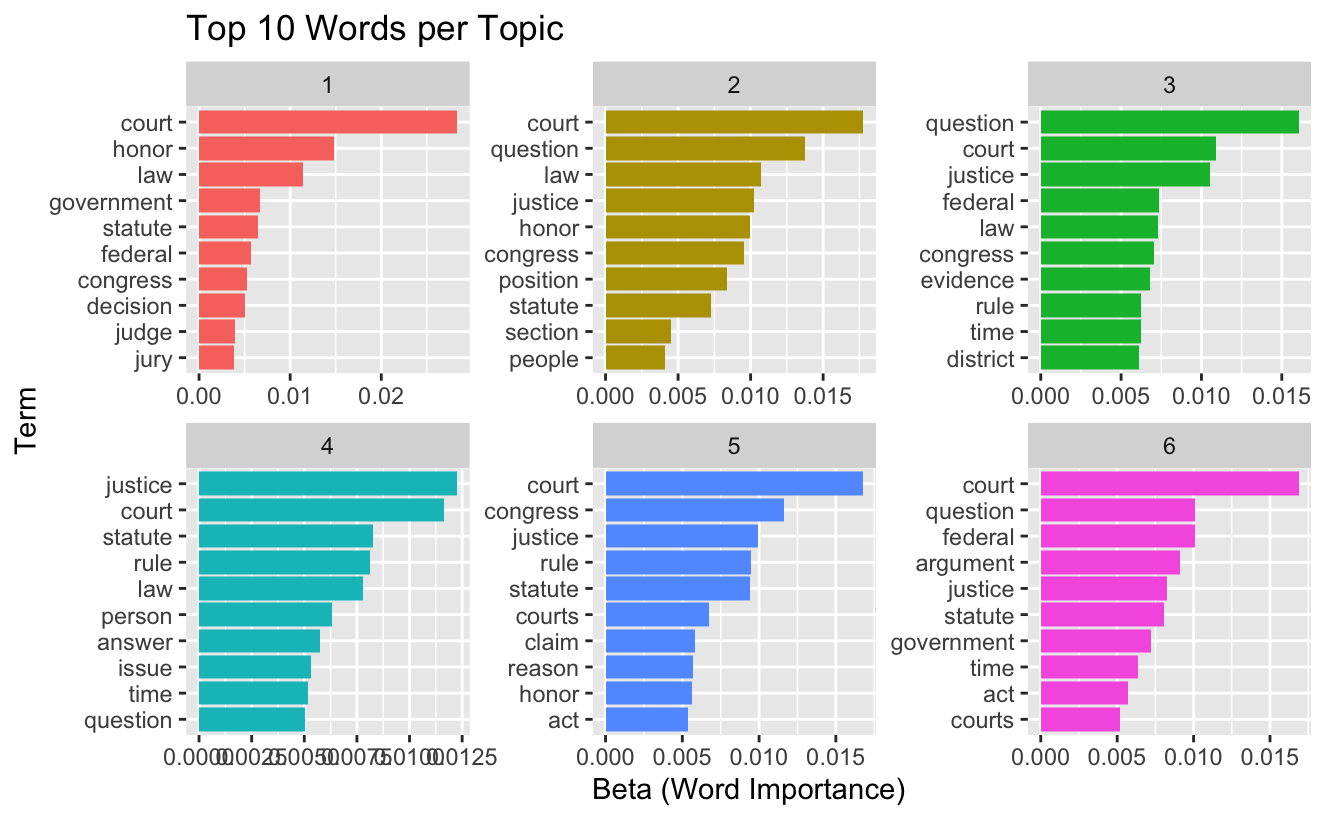}
    \caption{Top 10 terms for the six LDA topics}
    \label{fig:lda_topic_topterms}
\end{figure}

\end{document}